\documentclass{elsarticle}
\usepackage{lipsum}
\makeatletter
\def\ps@pprintTitle{%
 \let\@oddhead\@empty
 \let\@evenhead\@empty
 \def\@oddfoot{}%
 \let\@evenfoot\@oddfoot}
\makeatother

\journal{}

\usepackage{graphicx}
\usepackage{times}
\usepackage{epsfig}
\usepackage{graphicx}
\usepackage{amsmath}
\usepackage{amssymb}
\usepackage{pifont}
\usepackage{float}
\usepackage{xcolor}
\usepackage{bm}
\usepackage{times}
\usepackage{epsfig}
\usepackage{graphicx}
\usepackage{float}
\floatstyle{plaintop}
\restylefloat{table}
\usepackage{multirow}
\newcommand{\xmark}{\ding{55}}%
\newcommand{\cmark}{\ding{51}}%
\usepackage{array}
\newcolumntype{x}[1]{>{\centering\arraybackslash\hspace{0pt}}p{#1}}
\usepackage{subcaption}
\usepackage[symbol]{footmisc}
\begin{document}

\begin{frontmatter}

\title{LIAAD: Lightweight Attentive Angular Distillation for Large-scale Age-Invariant Face Recognition}

%\title{Beyond Disentangled Representations: An Attentive Angular Distillation Approach to Large-scale Lightweight Age-Invariant Face Recognition}

% \tnotetext[mytitlenote]{Fully documented templates are available in the elsarticle package on \href{http://www.ctan.org/tex-archive/macros/latex/contrib/elsarticle}{CTAN}.}

% \address{Radarweg 29, Amsterdam}
% \fntext[myfootnote]{Since 1880.}

\author{Thanh-Dat Truong$^1$\footnote[2]{Corresponding Author}}
\author{Chi Nhan Duong$^2$}
\author{Kha Gia Quach$^2$}
%\author{Dung Nguyen$^3$}
% \author{}
% \author{}
\author{Ngan~Le$^1$,~Tien~D.~Bui$^2$,~Khoa Luu$^1$}

\address{\small $^1$CVIU Lab, University of Arkansas, Fayetteville, USA}
\address{\small $^2$Department of Computer Science and Software Engineering, Concordia University, Canada}
%\address{$^3$VinAI, Vietnam}
\address{\texttt{\{tt032, khoaluu, thile\}@uark.edu, \{dcnhan, kquach\}@ieee.org, %v.dungnt244@vinai.io,
bui@encs.concordia.ca}}

% \author[mymainaddress,mysecondaryaddress]{Elsevier Inc}

% \author[mysecondaryaddress]{Global Customer Service\corref{mycorrespondingauthor}}
% \cortext[mycorrespondingauthor]{Corresponding author}

% \address[mymainaddress]{1600 John F Kennedy Boulevard, Philadelphia}
% \address[mysecondaryaddress]{360 Park Avenue South, New York}

\begin{abstract}
Disentangled representations have been commonly adopted to Age-invariant Face Recognition (AiFR) tasks. 
However, these methods have reached some limitations with (1) the requirement of large-scale face recognition (FR) training data with age labels, which is limited in practice; (2) heavy deep network architectures for high performance; and (3) their evaluations are usually taken place on age-related face databases while neglecting the standard large-scale FR databases to guarantee robustness. This work presents a novel Lightweight Attentive Angular Distillation (LIAAD) approach to Large-scale Lightweight AiFR that overcomes these limitations. Given two high-performance heavy networks as teachers with different specialized knowledge, LIAAD introduces a learning paradigm to efficiently distill the age-invariant attentive and angular knowledge from those teachers to a lightweight student network making it more powerful with higher FR accuracy and robust against age factor. Consequently, LIAAD approach is able to take the advantages of both FR datasets with and without age labels to train an AiFR model. Far apart from prior distillation methods mainly focusing on accuracy and compression ratios in closed-set problems, our LIAAD aims to solve the open-set problem, i.e. large-scale face recognition. Evaluations on LFW, IJB-B and IJB-C Janus, AgeDB and MegaFace-FGNet with one million distractors have demonstrated the efficiency of the proposed approach on light-weight structure.  This work also presents a new longitudinal face aging (LogiFace) database \footnote{This database will be made available} for further studies in age-related facial problems in future.
\end{abstract}

\begin{keyword}
Age-Invariant Face Recognition, Large-scale Face Recognition, Lightweight Network, Attentive Angular Distillation, Teacher-Student Network
\end{keyword}

\end{frontmatter}

% \linenumbers

% \footnotetext{Corresponding Author}

%\vspace{-5mm}
%%%%%%%%% BODY TEXT
\section{Introduction}
%\begin{figure}[t]
%	\centering \includegraphics[width=1\columnwidth]{Figs/peri_heatmap.png}
%	\caption{\textbf{Deep Attention Heat-map on Age-invariant Facial Regions.}}.
%	\label{fig:Peri_heatmap}
%\end{figure}

The research in Age-invariant Face Recognition (AiFR) has gained considerable prominence lately due to the challenges in the nature of human aging and the demand of consistent face recognition algorithms across ages. Indeed, such AiFR algorithms \cite{Xu_TIP2015} are important in practical applications where there is a significant age difference between probe and gallery facial photos, such as passport verification or missing children identification \cite{Luu_CAI2011}. However, compared to the state-of-the-art (SOTA) results of stand-alone Face Recognition (FR) algorithms, the performance of AiFR is still very limited due to the lack of robustly identifiable features stable across ages \cite{Luu_ROBUST2008}.
In addition, these AiFR algorithms are often evaluated separately from stand-alone FR algorithms although they are used together in practice. 

%\subsection{Motivation}
%\label{subsec:moti}
Disentangled learning representations have been widely used in AiFRs
% \cite{wen2016discriminative, liu2017sphereface, liu2018learning, sun2014deep, Wang_2018_CVPR, zhang2017range, schroff2015facenet, Wang_2018_CVPR, deng2018arcface}. 
% \cite{HD_LBP, HFA, MEFA, periocular_KhoaLuu, 7420684, DAL, OECNN, CAN}.
\cite{periocular_KhoaLuu, 7420684, DAL, OECNN, CAN}.
They however have reached some limitations. 
Firstly, in order to achieve a high accuracy performance, these AiFR methods often adopt heavy deep network architectures with the support of GPU platforms.
Then, they require a large-scale FR training set (i.e. multiple images per subject) with \textit{manual age labels}. However, this type of dataset is very limited in real-world. 
%Firstly, these methods often allow only
%face training sets with 
%
Indeed, there are many large-scale training face databases without age labels in practice \cite{guo2016ms, kemelmacher2016megaface}, but unusable for training these AiFR models. 
These disentangled learning based AiFR methods usually assume that the relationship between the identification and the age attributes can be linearly factorized in a latent or deep feature space.
% \cite{XX}.
Furthermore, these prior AiFRs are usually evaluated against age-related face databases, e.g. 
CACD-VS \cite{CACD_VS}, FG-Net \cite{fgnet_database}, IJB-B \cite{whitelam2017iarpa}, IJB-C \cite{maze2018iarpa}, AgeDB \cite{moschoglou2017agedb}, CA-LFW \cite{CA_LFW}, 
% \cite{CACD_VS, fgnet_database, maze2018iarpa, moschoglou2017agedb, whitelam2017iarpa, CA_LFW},
and have not been compared against other standard large-scale AiFR benchmarks \cite{sengupta2016frontal} to guarantee the robustness of the algorithms.

\begin{figure}[!t]
	\centering \includegraphics[width=0.95\columnwidth]{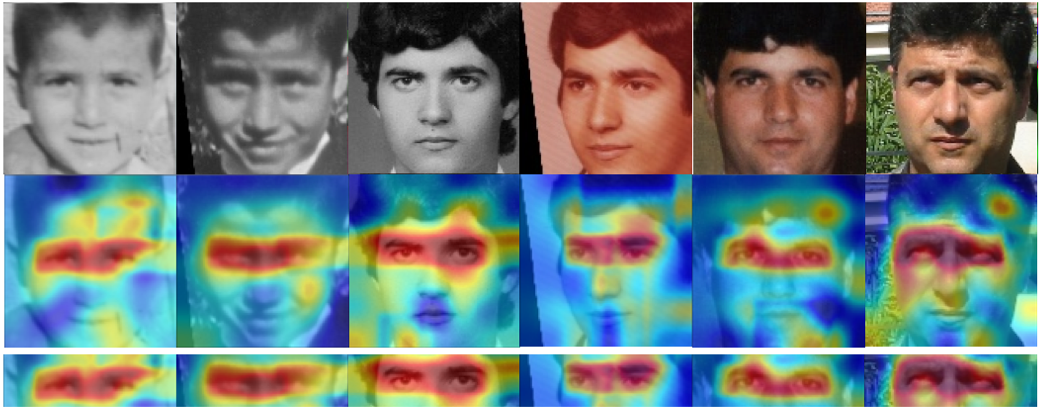}
	\caption{Deep Attention Heat-maps on Age-invariant regions.}
	\label{fig:Peri_heatmap}
% 	\vspace{-10mm}
\end{figure}

By addressing aforementioned limitations, this work presents a novel Lightweight Attentive Angular Distillation (LIAAD) approach to Large-scale Lightweight AiFR. Particularly, in order to alleviate the heaviness of a deep network structure while maintaining its accuracy, a Knowledge Distillation framework with two teachers, i.e. heavy high-performance networks, of different specialized knowledge is introduced. One teacher masters the FR task while the other tackles the age estimation task. Then, the attention knowledge about age-invariant facial regions (as shown in Fig. \ref{fig:Peri_heatmap}) and %\textcolor{red}
{feature discriminative power} from these teachers are distilled to the student via the proposed LIAAD. 
Consequently, the lightweight student network can naturally and effectively benefit from the knowledge of both teachers and becomes more powerful in both tasks.
Intuitively, with the knowledge from age estimation teacher, the student is guided to focus on the facial regions that are robust against age changes, while it is taught by the other teacher to achieve high accuracy on FR task. Then by generalizing these knowledge during training process, the student can be further improved for AiFR task. 

%\noindent
\textbf{Contributions. } This paper introduces an LIAAD framework for AiFR. The contributions of this work are five-fold.
(1) We proposed a novel \textit{Age-Invariant Attentive Distillation and Angular Distillation Losses} for distilling both age-invariant attention and feature direction from various teachers to the student. Since the teachers for different tasks are efficiently trained by the databases of those specific tasks, our LIAAD framework is able to take the \textit{advantages of both} face recognition datasets with and without age labels to train the models to generalize million-scale subjects. 
(2) Although angular metric has been recently used quite successfully in face recognition, it hasn't been discovered in network distillation problems. Our method looks at the Angular under a new point of view when it is used to translate the learned knowledge.
(3) Unlike prior distillation methods that mainly focus on closed-set problems with one or several teachers of the same task, our proposed distillation solution is proved to be even robust in the \textit{open-set problems}, i.e. large-scale face recognition. 
(4) The proposed LIAAD approach not only achieves state-of-the-art performance on AiFR databases but also is highly competitive against state-of-the-art face recognition methods on the standard large-scale face recognition benchmarks.
(5) Finally, this work introduces a new longitudinal face aging (LogiFace) database that will be made publicly available for further studies in age-related facial problems in future.
To the best of our knowledge, this is one of the first AiFR approaches that allow the use of training face databases with and without age labels together. Table \ref{tb:DistilledMethodReview} summarizes the difference between our proposed approach and the prior methods.

\begin{table*}[t]
	%\footnotesize
	\centering
	\caption{Comparison of the properties between our distillation approach and other methods. $\ell_{CE}$ denotes the cross-entropy loss. 
	%Feature Distribution (Feature Distribution), Activation (Act), Gradient (Grad).
	} 
	\label{tb:DistilledMethodReview} 
	%\small
	\resizebox{1.0\textwidth}{!} {
	\begin{tabular}{|l |c |c c| c c c|}
% 		\Xhline{2\arrayrulewidth}
        \hline
		\textbf{Method}  &
		\begin{tabular}{@{}c@{}}\textbf{Object }\\ \textbf{Class}\end{tabular}& \begin{tabular}{@{}c@{}}\textbf{Teacher}\\ \textbf{Transform}\end{tabular} & \begin{tabular}{@{}c@{}}\textbf{Student}\\ \textbf{Transform}\end{tabular}& \begin{tabular}{@{}c@{}}\textbf{Distilled}\\ \textbf{Knowledge}\end{tabular}& \begin{tabular}{@{}c@{}}\textbf{Loss}\\ \textbf{Function}\end{tabular}& \begin{tabular}{@{}c@{}}\textbf{Missing}\\ \textbf{Info.}\end{tabular}\\
		\hline
		KD \cite{Hinton_2015_NIPS} & Closed-set& Identity & Identity & Logits & $\ell_{CE}$ & \xmark\\
		FitNets \cite{Adriana_2015_Fitnets} & Closed-set& Identity & $1 \times 1$ conv & Magnitude& $\ell_2$ & \xmark\\
		%Attention Transfer
		Att\_Trans\cite{Zagoruyko_2017_AT} & Closed-set & Attention & Attention & Activation/Gradient Map& $\ell_2$ & \cmark\\
		DNN\_FSP \cite{Yim_2017_CVPR} & Closed-set & Correlation & Correlation &  Feat. Flow& $\ell_2$ & \cmark\\
		Jacob\_Match
		%Jacobian Matching 
		\cite{Srinivas_2018_Jacobian} & Closed-set & Gradient & Gradient & Jacobians& $\ell_2$ & \cmark\\
		 Factor\_Trans
		 %Factor Transfer 
		 \cite{Kim_2018_FT} & Closed-set & Encoder & Encoder & Feat. Factors& $\ell_1$ & \cmark\\
		Activation\_Bound
		%Activation Boundaries 
		\cite{Heo_2018_AB} & Closed-set & Binarization & $1 \times 1$ conv & Activation Map& Marginal $\ell_2$ & \cmark\\
		AL\_PSN\cite{wang2018adversarial} & Closed-set & Identity & Identity & Feature Distribution& $\ell_{GAN}$ & \cmark\\ 
		Robust SNL\cite{Guo_2018_CVPR} & Closed-set & Identity & Identity & Scores + Gradient& $\ell_2$ & \xmark\\ 
		\hline
		\hline
		\textbf{Ours} & \textbf{Open-set} & \textbf{Identity} & $\boldsymbol{1 \times 1}$ \textbf{conv} & \textbf{ Direction} & \textbf{Angular} & \xmark\\
% 		\Xhline{2\arrayrulewidth}
        \hline
	\end{tabular}
	}
	%\vspace{-5mm}
\end{table*}

\section{Related Work}

\subsection{Age Invariant Face Recognition}

Many previous works 
% \cite{HD_LBP, MEFA, HFA, periocular_KhoaLuu, 7420684, 5771107, DAL, OECNN, CAN, duong2019shrinkteanet, duong2018mobiface} 
\cite{periocular_KhoaLuu, DAL, OECNN, CAN, HFA, LFCNNs} 
explored the face invariant features from hand-craft features designed by heuristic to deep features learned by deep neural networks. Juefei-Xu et. al. \cite{periocular_KhoaLuu} presented a framework which utilizes the periocular region for age invariant face recognition.
% Gong et. al. \cite{HFA} proposed a Hidden Factor Analysis (HFA) capturing an identity factor that is age-invariant and an age factor affected by the aging process. The EM procedure was adopoted to estimates the latent factors .
Gong et. al. \cite{HFA} introduced a Hidden Factor Analysis (HFA) approach to decompose the latent face representation into age-invariant and age-sensitive latent factors. The process is optimized by using EM frameworks.
%The EM procedure was adopoted to estimates the latent factors .
Yandong et. al. \cite{LFCNNs} later introduced and improved version of HFA by using the deep convolutional neural network guided by Latent Factorization instead of the EM algorithm. Xu et. al. \cite{CAN} developed a coupled Auto-Encoder and a non-linear analysis factorizing identity feature from the face representation.
% proposed a method that learn age invariant features by the coupled Auto-Encoder and 
% presented a nonlinear factor analysis which uses to separate identity feature from face representation.
% Xu leveraged two shallow neural networks to connect two autoencoders which fit complex nonlinear aging and de-aging process.
Wang et. al. \cite{OECNN} proposed Orthogonal Embedding CNNs (OE-CNNs) decomposing deep face features into two orthogonal ID and Age components based on the directions in angular and radial spaces.
%where ID and age features are represented by angular and radial direction, respectively. 
% to represent age-related and identity-related features which helps to reduce the intraclass discrepancy caused by the aging. 
Wang et. al. \cite{DAL} proposed Decorrelated Adversarial Learning (DAL) algorithm utilizing the Batch Canonical Mapping Module to find the maximum correlation between the identity features and age features generated by a deep network. 
% correlation between the identity features and age features \cite{Duong_2017_ICCV, duong2019automatic, Chen_FG2011, Luu_BTAS2009} generated by a deep network. 

Zhao et. al. \cite{Zhao_PAMI2022} introduced a unified cross-age face synthesis and recognition framework to age-invariant face recognition. The entire network is effectively trained in an end-to-end manner to generate meaningful age-invariant facial representations distinguished from the age variations.
Similarly, Huang et. al. \cite{huang2020mtlface} presented a multi-task learning framework with an attention-based feature decomposition mechanism to jointly solve both age-invariant face recognition and face age synthesis. 
Meng et. al. \cite{meng2020multifeat} proposed a novel multi-features fusion and decomposition framework to learn deep feature representations and reduce the intra-class variations.
Moustafa et. al. \cite{Moustafa2020AgeinvariantFR} analyzed the potential of the deep features produced by VGG-Face \cite{Parkhi15} and discriminate the age-invariant related features using Multi Discriminant Correlation Analysis.
Tripathi et. al. \cite{TRIPATHI2021114786} proposed a novel local feature descriptor applied to parts-periocular region on faces to analyze the different pattern and dual directional relation pattern for AiFR. 
% a novel local feature descriptor to find difference pattern and dual directional relation pattern for age invariant face recognition. The proposed descriptor is applied over the preprocessed face images and its parts-periocular region i.e. left and right eye, mouth and nose region of a face image. 

\subsection{Knowledge Distillation} 

Many nature-inspired algorithms \cite{Hinton_2015_NIPS, Adriana_2015_Fitnets, Ebola_Optimization, Dwarf_Mongoose, Aquila_Optimizer, Reptile_Search} have been introduced to improve the learning capability of deep neural networks. 
Knowledge distillation \cite{Hinton_2015_NIPS, Adriana_2015_Fitnets} has emerged as an effective approach to learn lightweight deep neural networks. Rather than trying to ``simplify'' the computationally expensive deep network as in previous group, the Knowledge Distillation approaches in this direction aim at learning a light-weight network, i.e. the student, such that it can mimic the behaviors of the heavy one, i.e. the teacher. With the useful information from the teacher, the student can learn more efficient and be more ``intelligent''. 
Inspired by this motivation, one of the first knowledge distillation work was introduced by \cite{Ba_2013} suggesting to minimize the $\ell_2$ distance between the extracted features from the last layers of these two networks. Hilton et al. \cite{Hinton_2015_NIPS} later pointed out that the hidden relationships between the predicted class probabilities from the teacher are also very important and informative for the student. Then, the soft labels generated by the teacher model are adopted as the supervision signal in addition to the regular labeled training data during the training phase. 
In addition to the soft labels as in \cite{Hinton_2015_NIPS}, Romero et al. \cite{Adriana_2015_Fitnets} bridged the middle layers of the student and teacher networks and adopted $\ell_2$ loss to further supervise output of the student.

Several other aspects and knowledge of the teacher network are also exploited in the literature including 
%transferring the 
Feature Activation Map \cite{Heo_2018_AB}, Feature Distribution \cite{wang2018adversarial}, Block Feature Flow \cite{Yim_2017_CVPR}, Activation-based and Gradient-based Attention Maps \cite{Zagoruyko_2017_AT}, Jacobians \cite{Srinivas_2018_Jacobian}, Unsupervised Feature Factors \cite{Kim_2018_FT}. Recently, Guo et al. \cite{Guo_2018_CVPR} proposed to distill both prediction scores and gradient maps to enhance the student's robustness against data perturbations. 
Other knowledge distillation methods \cite{Mirzadeh_2019, Wang_2018_NIPS, Tommaso_2018_ICML, Zhang_2018_CVPR} are also proposed for variety of learning tasks.
Parituclarly, Wang et. al. \cite{Wang_2018_NIPS} propose knowledge distillation framework with Generative Adversarial Networks. 
Mirzadeh et. al. \cite{Mirzadeh_2019} reduces the knowledge gap between teacher and student networks by introducing teacher assistant network. 
% Yim et. al. \cite{Yim_2017_CVPR} introduced a fast optimization framework to knowledge distillation by considering the distilled
% knowledge as solving the flow of solution procedure matrix.
Tommaso et. al. \cite{Tommaso_2018_ICML} proposed two novel distillation objectives, i.e. Confidence-Weighted by Teacher Max and and Dark Knowledge with Permuted Predictions.
Zhang et. al. \cite{Zhang_2018_CVPR} presented an online framework for knowledge distillation.
% Chen et. al. \cite{Chen_2017_NIPS} introduced a knowledge distillation framework to object detection.
Although these methods have achieved prominent results, most of them are proposed for \textit{closed-set problems} with one or more teachers of the same task. In our work, we propose to distill the information from two teachers of different tasks to make the student more powerful with the desired property.

% \end{comment}
% \section{Our Proposed LIAAD Method}
\section{Learning with Knowledge Distillation and Age-Invariant Attentive Distillation}

\begin{figure}[!t]
	\centering \includegraphics[width=1.0\columnwidth]{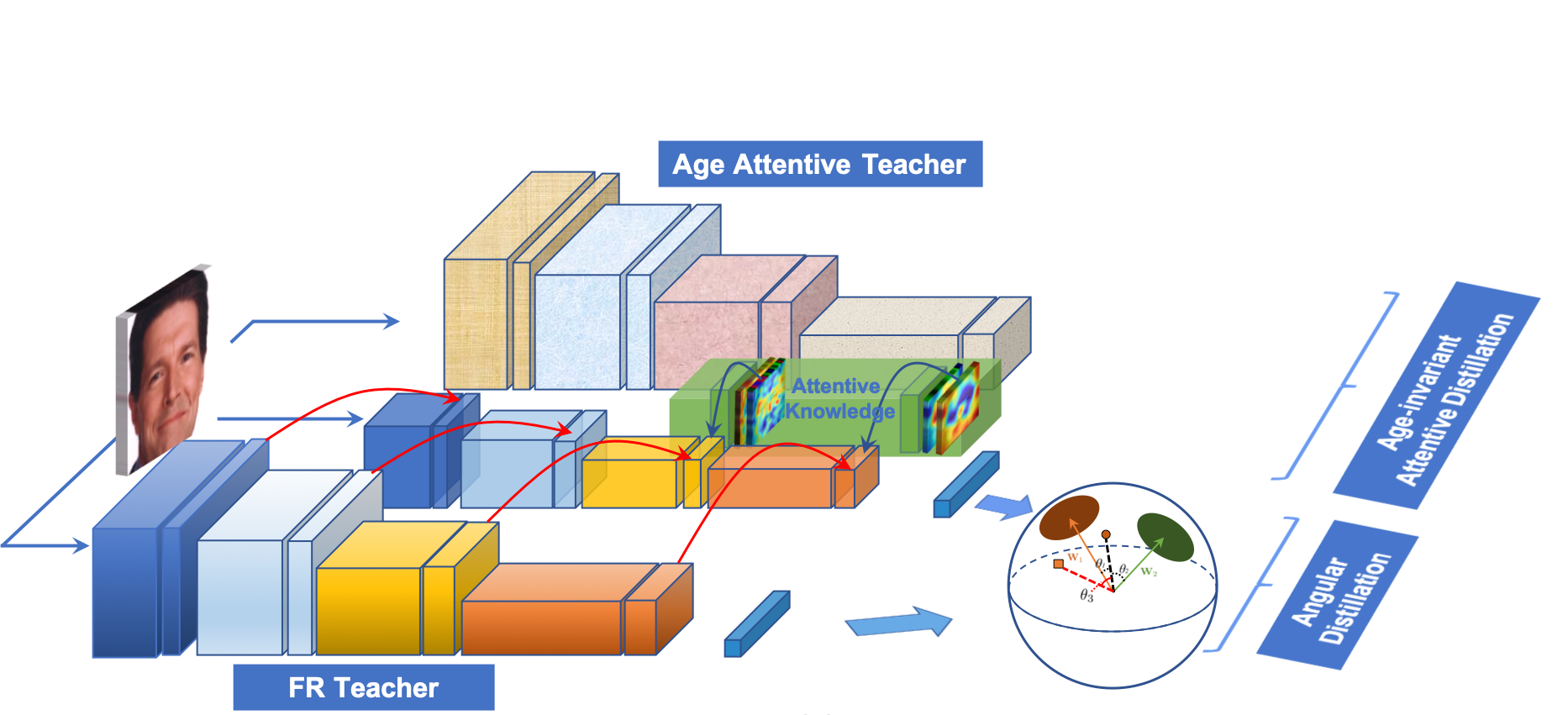}
	\caption{\textbf{The structures of our proposed LIAAD framework.} The top and the bottom networks illustrate the two teacher networks, i.e. the face recognition teacher (bottom figure), the age attentive teacher (top figure). The middle figure represents the student network that receives the distilled knowledge from both teacher networks. The red and blur arrows indicate the knowledge flow from the face recognition teacher and the age attentive teacher, respectively.
% 	The red arrow indicates the knowledge flow from the face recognition teacher and the blue arrow indicates the knowledge flow from the age attentive teacher. 
	The sphere represents the angular loss distillation illustration. 
	Different colors of network blocks illustrate different scale of feature maps.
}
% 	\vspace{-4mm}
	\label{fig:ProposedFramework}
\end{figure}

This section firstly describes the general form of the knowledge distillation problem. 
% 
% Then, two important design aspects for FR are considered including: (1) \textit{the representation of the distilled knowledge}; and (2) \textit{how to effectively transfer them between the teacher and the student}.
%In order to enable the age-invariant features
%In particular, to 
In order to enhance the student with age-invariant capability while preserving a high accuracy on standard FR, the age-invariant and age-sensitive knowledge is exploited. 
This knowledge is then transferred to the student through Age-invariant Attentive procedures as shown in Fig. \ref{fig:ProposedFramework}.
% This knowledge is then transferred to the student through Attentive and Angular Distillation procedures as shown in Fig.
% two types of knowledge including Age-invariant Attentive and Angular Knowledge are exploited from different teachers. 
% This knowledge is then transferred to the student through Attentive and Angular Distillation procedures as shown in Fig. \ref{fig:ProposedFramework}. %(a). %While the first 
In this way, the first component guides the student to pay more attention to facial regions robust against changes across ages, while the second component can teach and supervise its student to obtain the effective final solutions.

% \subsection{Problem Formulation}
\subsection{Learning with Knowledge Distillation}

Formally, let $\mathcal{T}^R,\mathcal{T}^A,\mathcal{S}: \mathcal{I} \mapsto \mathcal{Z}$ 
%and $\mathcal{T^A}: \mathcal{I} \mapsto \mathcal{Z}$ 
define the mapping functions from an image domain $\mathcal{I}$ to a  high-level embedding domain where $\mathcal{T}^R,\mathcal{T}^A$ are teacher networks and $\mathcal{S}$ is the student network. 
All functions $\mathcal{T}^R$,$\mathcal{T}^A$ and $\mathcal{S}$ are the compositions of $n$ sub-functions as.
%$\mathcal{T}^R_i$, $\mathcal{T}^A_i$ and $\mathcal{S}_i$ 
\begin{equation}
%\footnotesize
\begin{split}
    \mathcal{T}^R(I;\Theta^{t_r}) &= [\mathcal{T}^R_1 \circ \mathcal{T}^R_2 \circ \cdots \circ \mathcal{T}^R_n](I,\Theta^{t_r}) \\
    \mathcal{T}^A(I;\Theta^{t_a}) &= [\mathcal{T}^A_1 \circ \mathcal{T}^A_2 \circ \cdots \circ \mathcal{T}^A_n](I,\Theta^{t_a}) \\
    \mathcal{S}(I;\Theta^s) &= [\mathcal{S}_1 \circ \mathcal{S}_2 \circ \cdots \circ \mathcal{S}_n] (I,\Theta^s)
\end{split}
\end{equation}
where $I$ denotes the input image, $\circ$ is the function composition notation (e.g $\mathcal{T}_1 \circ \mathcal{T}_2 (I) = \mathcal{T}_1(\mathcal{T}_2(I))$), and $\{\Theta^{t_r},\Theta^{t_a},\Theta^s\}$ are the parameters of $\{\mathcal{T}^R,\mathcal{T}^A,\mathcal{S}\}$, respectively. Then given two \textit{high-capacity functions} $\mathcal{T}^R$ and $\mathcal{T}^A$ (i.e. \textit{teachers}), the goal of model distillation is to distill the knowledge from $\mathcal{T}^R$ and $\mathcal{T}^A$ to a \textit{limited-capacity function} $\mathcal{S}$ (i.e. \textit{student}) so that $\mathcal{S}$ can embed a similar latent domain as $\mathcal{T}^R$ with an additional desired features of $\mathcal{T}^A$.
To achieve this goal, the learning process of $\mathcal{S}$ is usually taken place under the supervision of $\mathcal{T}^R$ and $\mathcal{T}^A$ by comparing their outputs via $\mathcal{L}_{distill} = \sum_i^n \lambda_i \mathcal{L}_i(\mathcal{S},\mathcal{T}^R,\mathcal{T}^A)$ where
%In order to measure the qualification of $\mathcal{S}$ at each sub-component, the difference between $\mathcal{S}$ and $\mathcal{T}$ can be estimated as
\begin{equation} \label{eqn:LossDistill}
%\footnotesize
\begin{split}
    \mathcal{L}_i(\mathcal{S},\mathcal{T}^R,\mathcal{T}^A) &= \sum_{\mathcal{T} \in \{\mathcal{T}^R,\mathcal{T}^A\}} d \left(\mathcal{G}^{\mathcal{T}}_i(F^{t}_i), \mathcal{G}^{s}_i(F^s_i; \mathcal{T});\mathcal{T} \right) \\ 
    F^t_i &= \left[\mathcal{T}_1 \circ \mathcal{T}_2 \circ \cdots \circ \mathcal{T}_i\right](I,\Theta^t), t \in \{t_r, t_a\}\\
    F^s_i &= \left[\mathcal{S}_1 \circ \mathcal{S}_2 \circ \cdots \circ \mathcal{S}_i\right](I,\Theta^s) 
\end{split}
\end{equation}
where $\mathcal{G}^\mathcal{T}_i(\cdot)$ and $\mathcal{G}^s_i(\cdot)$ are transformation functions of $\mathcal{T}$ and $\mathcal{S}$; these transformations make their corresponding embedded features comparable. $d(\cdot,\cdot)$ denotes the difference between these transformed features.
%Then by minimizing these differences $\mathcal{L}_{distill} = \sum_i^n \lambda_i \mathcal{L}_i(\mathcal{S},\mathcal{T}^R,\mathcal{T}^A)$, the knowledge from the two teachers can be transferred to the student $\mathcal{S}$ so that they can embed similar latent domain with the desired features.
It is worth to note that the form of $\mathcal{L}_i(\mathcal{S},\mathcal{T}^R,\mathcal{T}^A)$ 
%in Eqn. (\ref{eqn:LossDistill}) 
provides two important properties. Firstly, since $d(\cdot,\cdot)$ measures the distance between $F_i^t$ and $F_i^s$, it implicitly defines the knowledge to be transferred from $\mathcal{T}$ to $\mathcal{S}$. Secondly, 
%the transformation functions 
$\mathcal{G}^t_i(\cdot)$ and $\mathcal{G}^s_i(\cdot)$ control the portion of the transferred information.
The next sections focus on the designs of these two components for selecting the most informative features and transferring them to the student. 

% \begin{figure}[!t]
% 	\centering \includegraphics[width=1.0\columnwidth]{Figs/ProposedFramework_new1.png}
% 	\caption{The structures of (a) our Proposed LIAAD Framework and (b) Age-Attention Block to extract Age-Invariant Attentive Knowledge from the teacher.}
% % 	\vspace{-4mm}
% 	\label{fig:ProposedFramework}
% \end{figure}

\subsection{Age-invariant Attention from Teacher}

Generally, among the informative knowledge that can be exploited from the teacher, the attention mechanism 
% \cite{Zagoruyko_2017_AT, ADL_CITE, CAM_CMU, Singh_2017_ICCV, Wei_2017_CVPR} 
\cite{Zagoruyko_2017_AT, ADL_CITE, CAM_CMU} 
to make decisions is one of the key aspects to enhance the power of a teacher as well as to improve the knowledge of student $\mathcal{S}$. 
%For example, an object detection teacher can provide its student more ideas about the effective regions to find an object; or a object recognition teacher can show which features should be focused to classify them. %
This is also consistent with our visual experience where attention information is important for human to possess details and coherence. 
Therefore, in this section, we propose an attentive distillation process making the student $\mathcal{S}$  enhanced with desired properties, especially the age-invariant factor.
Particularly, to obtain the age-invariant property for $\mathcal{S}$, in the simplest approach, one can derive a teacher function with that capability (i.e. \textit{map the input image $I$ to an age-invariant latent feature space}) and exploit the attention information from that teacher. However, obtaining this teacher function is not an easy task.

Although some previous disentangled learning approaches have been introduced in literature \cite{DAL, OECNN, HFA}
% [HFA, OE-CNNs, DAL]
, most of them rely on the assumption that ID and age attributes can be linearly factorized in embedding domain. Moreover, the training data for these approaches are still limited.
Meanwhile, we argue that this goal can still be practically achieved from an alternative teacher who can show its student about age-sensitive regions of the face. From that instruction, the student can generalize its knowledge by adaptively paying more attention to the remaining face regions to reduce the sensitivity to age factor.
%
%or an age estimation teacher can effectively teach its student about age-sensitive regions of the face. 
%
%Formally, let $\mathcal{T}^{A}$ be a teacher function with a desired property. Without loss of generality, we assume that $\mathcal{T}^{A}$ is an age-estimation function that maps the input image $I$ to the age of a subject in $I$.
%
%
%\noindent
%\textbf{Attention Knowledge.} 

\begin{figure}[!t]
	\centering \includegraphics[width=0.65\columnwidth]{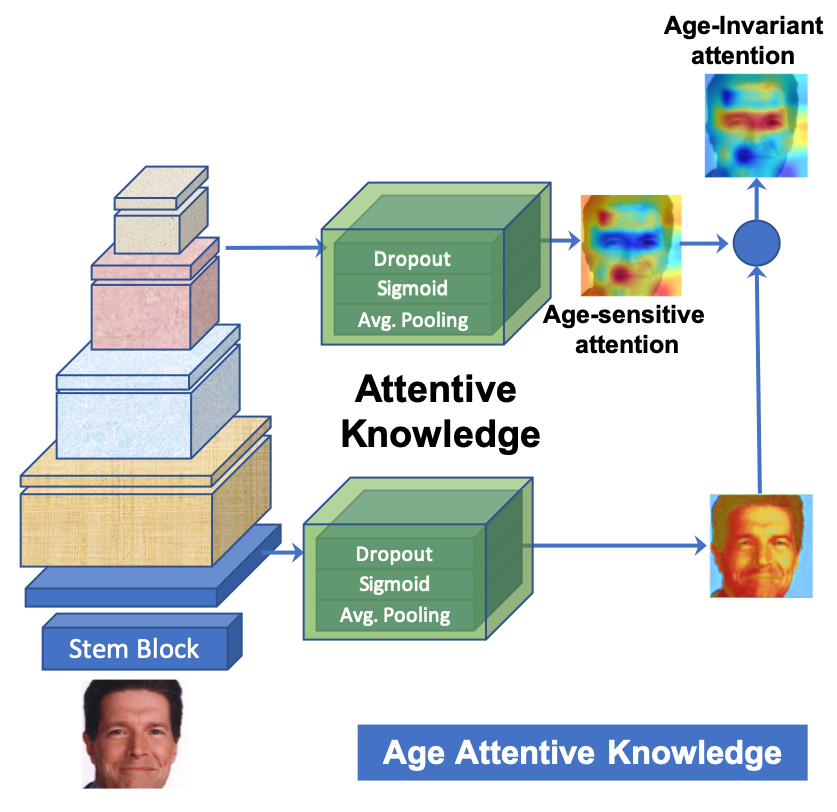}
	\caption{Age-Attention Block to extract Age-Invariant Attentive Knowledge from the teacher.}
% 	\vspace{-4mm}
	\label{fig:ProposedFramework_b}
\end{figure}

\begin{figure}[!b]
    \centering
    \includegraphics[width=0.8\columnwidth]{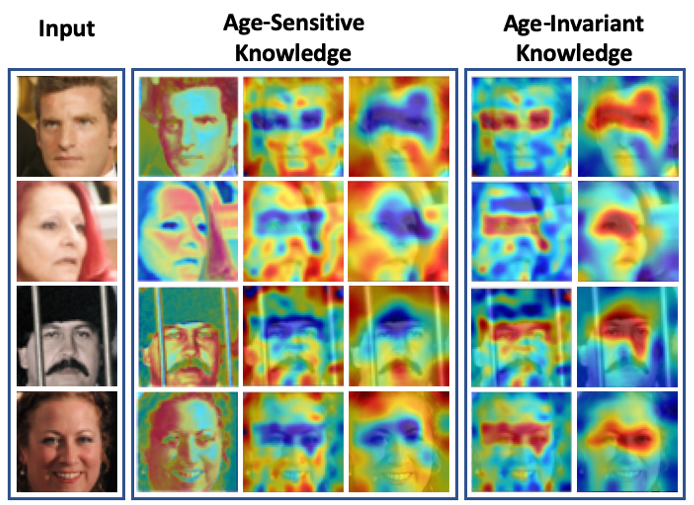}
    \caption{\textbf{Attentive Knowledge from $\mathcal{T}^A$ with ResNet-18 structure.} Given an input (the 1st column), the age-sensitive attentive knowledge (the 2nd to the 4th columns) at different levels according to the depth of ResNet-unit
    are extracted. Then the knowledges at higher levels are adaptively inverted for age-invariant attentive knowledge (the 5th to the 6th columns).}
    %Blended attention heat map of Age Estimation model. The first group is input images, the second group is the blended age-sensitive heat map of stem block and two layers in the last block of age estimation model, the last group is the blended the age-invariant attention heat map. [THIS FIGURE WILL BE REPLACED]}
    \label{fig:AttentionMap}
    % \vspace{-10mm}
\end{figure}

%\noindent
\textbf{Age-Sensitive Attentive Knowledge.} Formally, let $\mathcal{T}^{A}$ be a teacher function mapping the input image $I$ to the latent features used to predict the age of a subject in $I$. 
%By investigating the behaviour of $\mathcal{T}^{A}$, we found that it does not process equally on an input feature map, it focus on some important regions of the input feature map which are sensitive with facial aging. 
%This phenomenon is called \textit{attention mechanism} raised in recent years \cite{ADL_CITE, CAM_CMU, Wei_2017_CVPR, Singh_2017_ICCV}.
An attentive mapping function $\mathcal{A}^t_i(\cdot)$ with respect to a feature map $F^{t_a}_i$ can be defined as $\mathcal{A}^t_i: \mathbb{R}^{c_i \times h_i \times w_i} \mapsto \mathbb{R}^{h_i \times w_i}$ where each value in $A^t_i$
indicates the importance of the corresponding entry in $F^{t_a}_i$. In other words, the function $\mathcal{A}^t_i$ aims to validate the contributions of each local region in $F^{t_a}_i$ to the discriminative power of the final representation of $\mathcal{T}^{A}$. 
Then by looking at the regions with higher importance, we can efficiently extract the knowledge of where $\mathcal{T}^{A}$ pays attention in each stage of its embedding process. Intuitively, since $\mathcal{T}^{A}$ aims to discriminate the age factor presented in $I$, its embedding process will naturally focus on the age-sensitive facial regions and automatically enhances the importance of these regions with higher intensity values for $F^{t_a}_i$. 
%the more important regions  
This type of information can be considered as the age-sensitive attentive knowledge and extracted via the statistics across $c_i$ channels of $F^{t_a}_i$.
%
%Since $F^{t\_a}_i$ represents an intermediate features during embedding process of a classification task, the intensity of each entry in $F^{t\_a}_i$ is proportional to its contribution to the discriminative power of the final representation. Therefore,  
%the attentive mapping function $\mathcal{A}^t_i$ can be defined based on the statistics across $c_i$ channels of $F^{t\_a}_i$.
For example, if $F^{t_a}_i$ is a feature map extracted from a Deep Neural Network (DNN), an average pooling operator followed by an activation function can be effectively adopted for $\mathcal{A}^t_i$. 
%Then it can be converted to the an attention map as
\begin{equation}
%\footnotesize
    A^T_i = \sigma(\mathcal{A}^t_i(F^{t_a}_i))
\end{equation}
where $\sigma(\cdot)$ denotes an activation function in which its output goes into the value range to $[0,1]$.
In our experiment, we choose Sigmoid as our activation function $\sigma(\cdot)$. 
It should be noted that any activation function satisfying our mentioned condition can be adopted.
There are two properties of $\mathcal{A}^t_i$. Firstly, a region with higher value of $A^t_i$ is more sensitive to age factor compared to the others.
%indicates the more age-sensitive region of $F^{t_a}_i$.
Secondly, according to the depth of $A^T_i$ in the embedding process, different levels of attentions can be extracted.
%each sub-function gives different level of attention according to the depth of that sub-function. 
Fig. \ref{fig:AttentionMap} illustrates different attention levels of ResNet-18 structure trained for age estimation task. The design of $A^T_i$ is shown in Fig. \ref{fig:ProposedFramework_b}
% (b) 
.where three operations are adopted including Average Pooling, Sigmoid activation and a dropout layer. This dropout layer can improve the quality of the learned attention feature map, and, therefore, enhance the capability of $A^i_i$ to discriminate between age-sensitive and age-invariant regions.
% to enhance the discriminative power for $\mathcal{T}^{A}$. 
Interestingly, the attention mask of the first sub-function (i.e. ResNet unit) gives a coarse attention to the whole facial region, while the last sub-function pays more attention to specific age-sensitive facial regions.

%\noindent
\textbf{Age-invariant Attentive Knowledge.}  
%the age-invariant face recognition model needs to concentrate on region which is invariant to facial aging.
Although the age-sensitive attentive knowledge can give some hints to the student $S$ to deal with the aging factor, this knowledge cannot be directly distilled to $S$. In fact, the goal of $S$ is to be robust against age changing. Therefore, a natural way for $S$ to benefit from $A^T_i$ is to pay attention to the ``inversion'' of $A^T_i$. In other words, we can ``flip'' the attention of $S$ to the regions that give little contribution to age estimator process, and, therefore, less sensitive to aging changes. Since all values of $A^T_i$ are within the range of $[0,1]$, an inverse flipping operator can be defined as $1 - A^T_i$. Normally, this operator can give an initial estimation of potential candidates for the age-invariant knowledge. However, the output of this flipping operator may still include other unrelated-regions such as the background outside the face. 
Interestingly, as illustrated in Fig. \ref{fig:AttentionMap}, although the attention map at a coarsest level of $A^T_1$ provides very low-level attention with more focus on face regions while disregarding the ones outside the face. 
Therefore, the age-invariant attentive knowledge for $S$ can be effectively extracted by incorporating this knowledge to the inverse flipping of $A^T_i$.
\begin{equation}
%\footnotesize
    \hat{A}^T_i = \mathcal{D}_i(A^T_1) \odot (1 - A^T_i), i > 1
\end{equation}
where $\odot$ represents the Hadamard product; and $\mathcal{D}_i(\cdot)$ denotes a down-sampling operator to match the dimensions between $\mathcal{D}_i(A^T_1)$ and $A^T_i$.
The last column of Fig. \ref{fig:AttentionMap} illustrates the age-invariant attention map obtained by $\hat{A}^T_i$ extracted from ResNet structure. Interestingly, the age-invariant attention is mainly distributed around the eyes of the faces. This is consistent with the finding from \cite{periocular_KhoaLuu}. However, different from that work, rather than manually pre-defining the periocular region, the design of $\hat{A}^T_i$ can help to adaptively extract them from the input. Moreover, this design becomes more robust and consistent against poses. 

%\noindent
\subsection{Age-invariant Attentive Distillation}
% \textbf{Age-invariant Attentive Distillation.} 
Finally, the knowledge from $\hat{A}^T_i$ now can be effectively distilled to the student $\mathcal{S}$ as the age-invariant attentive distillation loss is defined as follows: 
% by
%Then this attentive knowledge can be distilled to the student $S$ by
\begin{equation} \label{eqn:AttentionDistill}
%\footnotesize
\begin{split}
    \mathcal{L}^{Att}(\mathcal{S}, \mathcal{T}^A)&= \sum_{i>1} %\mathcal{L}_i^A(\mathcal{S}, \mathcal{T}^A) =
    || \mathcal{G}_i^{t_a}(\hat{A}^T_i) -\mathcal{G}_i^{s_a}(A^S_i)||_2^2
\end{split}
\end{equation}
where $\mathcal{G}_i^{t_a}$ and $\mathcal{G}_i^{s_a}$ are the transformation functions matching the spatial dimensions of $\hat{A}^T_i$ and $A^S_i$. Generally, we can choose the corresponding of $\hat{A}^T_i$ and $A^S_i$ in the structure of $\mathcal{T}^A$ and $\mathcal{S}$ and choose the identity transformation for $\mathcal{G}_i^{t_a}$ and $\mathcal{G}_i^{s_a}$ to prevent missing information during distillation. %Other choices for these functions are still applicable. 
During distillation process, to further enhance the attention flow through different levels during the embedding process of $\mathcal{S}$, the learned attention mask is also incorporated back to each feature $F^s_i$ by $F_i^s = A_i^S \otimes F_i^s$ %as shown in Fig. XXX 
where $\otimes$ denotes the spatial multiplication operator.

\section{Lightweight Attentive Angular Distillation}

% \subsection{Feature Direction Knowledge from Teacher}

In this section, we further analyze the effectiveness of the distillation process. In particular,
% Then, 
two important design aspects for FR are considered including: (1) \textit{the representation of the distilled knowledge}; and (2) \textit{how to effectively transfer them between the teacher and the student}. 
We firstly revise the standard knowledge distillation loss and the softmax loss in classification problem. Then, we further describe our proposed Angular Distillation Loss and the the proposed Lightweight Attentive Angular Distillation Network.
%In order to enable the age-invariant features
%In particular, to 

\subsection{Feature Direction Knowledge from Teacher}
As in Table \ref{tb:DistilledMethodReview}, most prior distillation frameworks are introduced in the closed-set classification problem, i.e. object classification or semantic segmentation with predefined classes. With the assumption about the fixed (and small) number of classes, traditional metrics can be efficiently adopted for the distillation process. 
For example, given the teacher $\mathcal{T}^R$, $\ell_2$-norm distance is usually applied to measure the similarity between $\mathcal{S}$ and $\mathcal{T}^R$, i.e. $d(\mathcal{G}^{t_r}_i(F^{t_r}_i), \mathcal{G}^{s_r}_i(F^s_i)) = \parallel \mathcal{G}^{t_r}_i(F^{t_r}_i) -  \mathcal{G}^{s_r}_i(F^{s}_i) \parallel^2_2$. Since the capacity of $\mathcal{S}$ is limited, employing this constraint as a regularization to each $F^s_i$ %(i.e. to enforce $F^s_i$ and $F^t_i$ to be exact matched)
%having similar magnitude and direction) 
can lead to the over-regularized issue. 
%Consequently, this constraint becomes too hard and makes the learning process more difficult.
Another metric is to adopt the class probabilities predicted from $\mathcal{T}^R$ as the soft target distribution for the student $\mathcal{S}$ \cite{Hinton_2015_NIPS}. However, this metric is efficient only when the object classes are fixed in both training and testing phases. Otherwise, the knowledge to convert embedded features to class probabilities cannot be reused during testing stage and, therefore, the distilled knowledge is partially ignored.

In open-set problems, 
%classification or verification problems such as face recognition or verification, 
since classes are not predefined beforehand, the sample distributions of each class and the margin between classes become more valuable knowledge. In other words, the angular differences between samples and how the samples distributed in the teacher's hypersphere are more beneficial to the student.
Therefore, we propose to use the angular information as the main knowledge to be distilled. 
By this way, rather than enforcing the student to follow the exact outputs of the teacher (as in case of $\ell_2$ distance), we can relax the constraint so that the embedded features extracted by the student only need to have similar direction as those extracted by the teacher.
Generally, with ``softer'' distillation constraint, the student is able to adaptively interpret the teacher's information and learns the solution process more efficiently. 

\subsection{Softmax Loss Revisit}
%\noindent
% \textbf{Softmax Loss Revisit.} 
As one of the most widely used losses for classification problem, Softmax loss of each input image is formulated as follows. 
%The most widely used classification loss function, softmax loss, of each input image is presented as follows:
\begin{equation}
\begin{split}
%\footnotesize
    \mathcal{L}_{SM} & = -\log \frac{e^{\mathbf{W}_y* F^s_n}}{\sum_{c} e^{\mathbf{W}_c * F^s_n} } \\
    & = -\log \frac{e^{\parallel\mathbf{W}_y\parallel \parallel F^s_n \parallel \cos \theta_y}}{\sum_{c=1}^C e^{\parallel \mathbf{W}_c\parallel \parallel F^s_n\parallel \cos \theta_c} } \label{eqn:SMLoss}
\end{split}
\end{equation}
where $y$ is the index of the correct class of the input image and $C$ denotes the number of classes. Notice that the bias term is fixed to $0$ for simplicity.
By adopting the $\ell_2$-norm normalisation to both feature $F^s_n$ and weight $\mathbf{W}_c$, the angle between them becomes the only classification criteria. If each weight vector $\mathbf{W}_c$ is regarded as the representative of class $c$, minimizing the loss means that the samples of each class are required to be distributed around that class' representative with the minimal angular difference.
%
%to classify examples correctly, the direction of a weight vector can be regarded as the representative to the normalized features of the corresponding class and the samples of that class are required to distributed around that representative with the minimal angular difference.
%
% This is also true during testing process where the angle between the direction of extracted features from input image and the representative of each class (i.e. classification problem with predefined classes) or  extracted features of other sample (i.e. verification problem) are used for deciding whether they belong to the same class. 
This is also true during the testing process where the angles are between the direction of extracted features from input image and the direction of representative of each class (i.e. classification problem with predefined classes) or the direction of extracted features of other sample (i.e. verification problem to decide whether the input and the sample belong to the same class).
In this respect, the magnitude of the feature $F^s_n$ becomes less important than its direction. Therefore, rather than considering both magnitude and direction of $F_n^t$ for a distillation process, knowledge about the direction is enough for the student to achieve similar distribution to the teacher's hypersphere. Moreover, this knowledge can be also efficiently reused to compare samples of object classes other than the ones in training.
%Therefore, considering the directions of teacher and student features, as long as $F^s_n$ and $F^t_n$ have similar direction, these features can freely distributed on different hyperspheres with various radius in latent space.

\subsection{Angular Distillation Knowledge}
\begin{figure}[!t]
	\centering \includegraphics[width=1.0\columnwidth]{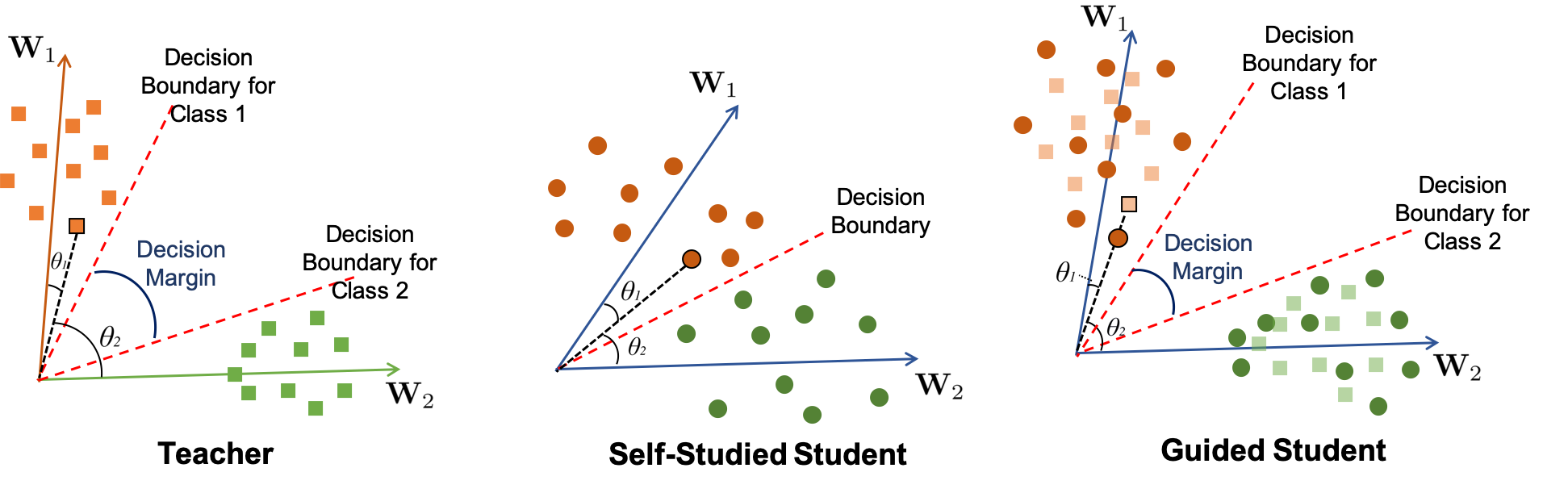}
	\caption{\textbf{Geometric Interpretation of Angular Distillation Loss.} 
	The square points and circle points refer to features produced by the teacher network and the student network, respectively.
	With high-capacity function $\mathcal{T}$, the teacher is able to produce a large decision margin between two classes while the self-studied student only gives small decision margin. By following the direction provided by the teacher, the student can make better decision boundaries with larger margin between classes. 
	%Moreover, with the angular distillation loss, the student is not strictly required to produce exact feature as its teacher (\textit{i.e. which is a very hard constraint according to the student's capability)}.
	}
	\label{fig:GeometricInterpretation}
	%\vspace{-5mm}
\end{figure}

%\noindent
% \textbf{Angular Distillation Knowledge.} %Inspired by this motivation, 
We propose to use the direction of the teacher feature $F^t_n$ as the distilled knowledge and define an \textbf{angular distillation loss} as follows.
\begin{equation}
%\footnotesize
    \mathcal{L}_n(\mathcal{S}, \mathcal{T}^R) 
    %= d(\mathcal{G}^t_n(F^t_n), \mathcal{G}^s_n(F^s_n)) 
    = \left\| 1 - \frac{\mathcal{G}^{t_r}_n(F^{t_r}_n)}{\parallel \mathcal{G}^{t_r}_n(F^{t_r}_n)\parallel}* \frac{\mathcal{G}^{s_r}_n(F^s_n)}{\parallel \mathcal{G}^{s_r}_n(F^s_n)\parallel}\right\| ^2_2
\end{equation}
With this form of distillation, 
%the only knowledge needed to be transferred between $\mathcal{T}$ and $\mathcal{S}$ is the direction of embedded features. In other words, 
as long as $F^s_n$ and $F^t_n$ have similar directions, %these features 
they can be freely distributed on different hyperspheres with various radius in latent space. This produces a degree of freedom for $\mathcal{S}$ to interpret its teacher's knowledge during learning process. 
Incorporating this distillation loss to Eqn. (\ref{eqn:SMLoss}), the objective function becomes.
\begin{equation} \label{eqn:TotalLoss}
%\footnotesize
    \mathcal{L} = \mathcal{L}_{SM} + \lambda_n \mathcal{L}_n(\mathcal{S}, \mathcal{T}^R)
\end{equation}
%
%In Eqn. (\ref{eqn:TotalLoss}), 
The first term corresponds to the traditional classification loss whereas the second term guides the student to learn from the teacher's hypersphere. Notice that, 
%this objective function is not limited to specific classification loss. 
this distillation loss can also act as a support to any other loss functions. 
To prevent missing information during distillation, we choose identity transformation for $\mathcal{T}^R$, while  $\mathcal{G}^s_n(F^s_n)$ is chosen as a mapping function such that the dimension of $F^s_n$ is increased to match the dimension of $F^t_n$, i.e. $1 \times 1$ convolution operator. By this way, no information is missing during feature transformation and, therefore, $\mathcal{S}$ can take full advantages of all knowledge from $\mathcal{T}$. Geometric interpretation of the angular distillation knowledge is illustrated in Fig.  \ref{fig:GeometricInterpretation} by considering the binary classification of two classes with the representatives $\mathbf{W}_1$ and $\mathbf{W}_2$. With higher capacity, the teacher can provide better decision margin between the two classes, while the self-studied student (i.e. using only softmax loss function) can only give small decision margin.
When the $\mathcal{L}_n(\mathcal{S}, \mathcal{T}^R)$ is incorporated (i.e. guided student), the classification margin between class 1 and class 2 is further enhanced by following the feature directions of its teacher and produces better decision boundaries. %. Figure \ref{fig:GeometricInterpretation} illustrates the decision boundaries when the student learns by itself using only softmax loss and learns with the guide from teacher. 
Furthermore, one can easily see that even when the student is not able to produce exact features as its teacher, it can easily mimic the teacher's feature directions and benefit from the teacher's hypersphere. 
%Fig. \ref{fig:ShrinkTeaNetLastLayer} illustrates the distillation process for Angular knowledge.

\begin{figure}[t]
	\centering \includegraphics[width=1.0\columnwidth]{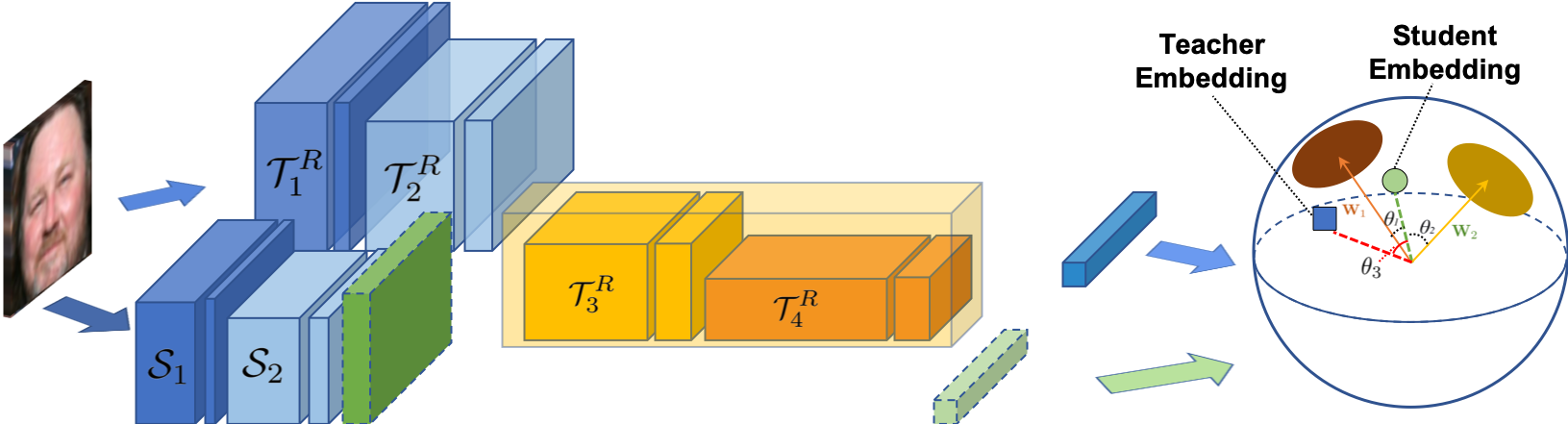}
	\caption{\textbf{Intermediate Angular Knowledge Distillation.} Given an intermediate student's features, i.e. $\mathcal{H}_i^s(F_i^s)$, the teacher uses its power to validate whether the student's input is informative enough to make similar decision as the teacher.}
	\label{fig:ShrinkTeaNetIntermedidate}
	%\vspace{-6mm}
\end{figure}

%\noindent
\textbf{Intermediate Angular Distillation Knowledge.}
%In this section, we further distill the knowledge of the teacher to intermediate components of the student. 
Generally, if the input to $\mathcal{S}$ is interpreted as the question and the distribution of its embedded features is its answer, the generated features at the middle stage, i.e. $F^s_i$, can be viewed as the 
%intermediate results 
intermediate understanding or interpretation of the student about the solution process. Then to help the student efficiently ``understand'' the development of the solution, the teacher should illustrate to the student \textit{``how the good features look like''} and \textit{``whether the current features of the student is good enough to get the solution in later steps''}. Then, the teacher can supervise and efficiently correct the student from the beginning and, therefore, leading to more efficient learning process of the student.
%
%Similar to previous section, rather than employing $\ell_2$ norm as the cost function of each pair $\{F^s_i,F^t_i\}$, 
Inspired by this motivation, we proposed to validate the quality of $F^s_i$ based on its angular difference between the embedding produced by $F^s_i$ and $F^{t_r}_i$ using the same teacher's interpretation toward the last stages. Particularly, given the intermediate feature $F^s_i$, it is firstly transformed by $\mathcal{H}^s_i$ to match the dimension of $F^{t_r}_i$. Then both $F^s_i$ and $F^{t_r}_i$ are analyzed by the teacher $\mathcal{T}^R$, i.e. $\tilde{\mathcal{T}}_i^R = \left[\mathcal{T}^R_{i+1}\circ \cdots \circ \mathcal{T}^R_n \right]$, for the final embedding features. Finally, their similarity in the hypersphere is used to validate the knowledge that $F^s_i$ embeds.
%In particular, the distillation loss for each intermediate feature $F^s_i$ can be formulated as follows.
\begin{equation} \label{eqn:DistillIntermediate}
%\footnotesize
\begin{split}
    \mathcal{L}_i(\mathcal{S}, \mathcal{T}^R) 
    %= & d(\mathcal{G}^t_i(F^t_i),\mathcal{G}^s_i(F^s_i))\\
    = & d(\tilde{\mathcal{T}}_i^R(F^{t_r}_i), \left[\mathcal{H}^s_i \circ \tilde{\mathcal{T}}_i^R \right](F^s_i))\\
    = & \left\| 1 - \frac{\tilde{\mathcal{T}}_i^R(F^{t_r}_i)}{\parallel \tilde{\mathcal{T}}_i^R(F^{t_r}_i)\parallel}* \frac{\left[\mathcal{H}^s_i \circ \tilde{\mathcal{T}}_i^R \right](F^s_i)}{\parallel \left[\mathcal{H}^s_i \circ \tilde{\mathcal{T}}_i^R \right](F^s_i)\parallel}\right\| ^2_2 
\end{split}
\end{equation}
The intuition behind this distillation loss is to validate whether $F^s_i$ contains enough useful information to make similar decision as its teacher in the later steps
%To validate this point, 
%to validate the understanding of the student at $i$-th stage, 
%we propose to take advantage of 
by borrowing the teacher power to solve the solution given the student input at $i$-th stage, i.e. $F^s_i$. In case the teacher still get similar solution with that input, the student's understanding until that stage is acceptable. Otherwise, the student is required to be re-corrected. 
Fig. \ref{fig:ShrinkTeaNetIntermedidate} illustrates the distillation for intermediate knowledge.

\subsection{Lightweight Attentive Angular Distillation Network}
%FIGURE TO SHOW THE NETWORK STRUCTURE FOR FACE RECOGNITION

Fig. \ref{fig:ProposedFramework} illustrates our proposed approach to distill both age-invariant attentive and angular distillation knowledge. ResNet-styled CNN with four ResNet blocks adopted for both teacher and student. The whole network can be considered as the mapping functions whereas each ResNet block corresponds to each sub-function.
%, i.e. $\mathcal{T}_i$ and $\mathcal{S}_i$.
It learns a strong but efficient student network by distilling the knowledge to all four blocks.
Given a dataset $D=\{I_j,y_j\}_{j=1}^N$ consisting of $N$ images $I_j$ and their corresponding labels $y_j$.
The overall learning objective is,
%the composition of Eqns. (\ref{eqn:TotalLoss}) and (\ref{eqn:DistillIntermediate}).
\begin{equation} \label{eqn:OverallLoss}
%\footnotesize
    \mathcal{L} = \frac{1}{N} \sum_j \left[ \mathcal{L}_{SM} + \lambda^A \mathcal{L}^{Att} +  \sum_i \lambda_i \mathcal{L}_i(\mathcal{S}, \mathcal{T}^R)\right]
\end{equation}
where $\lambda_i$ and $\lambda^A$ denote the hyper-parameters controlling the balance between the distilled knowledge to be transferred at different ResNet blocks. Moreover, the identity transformation function is used for all functions $G^t_i$ while $1 \times 1$ convolution followed by a batch normalization layer is adopted for $G^s_i$ to match the dimension of the corresponding teacher's features.

\section{Experimental Results}

This work is evaluated on the standard AiFR datasets and compared against the state-of-the-art methods. Unlike prior AiFR works, the proposed approach is also evaluated on the standard large-scale FR benchmarks and compared against recent light-weight FR methods. In addition, we also present a new longitudinal face aging (LogiFace) database for further studies in face-related problems in future.

\subsection{Longitudinal Face Aging (LogiFace) Database}

\begin{figure}[!t]
    \centering
    \includegraphics[width=1.0\columnwidth]{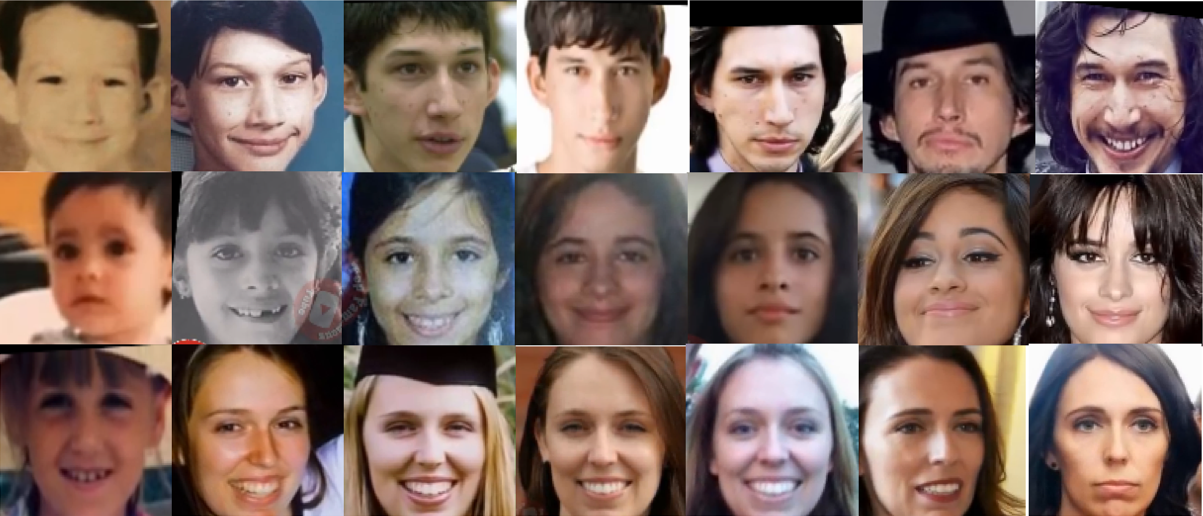}
    \caption{\textbf{Examples of Longitudinal LogiFace Database} where each row presents the images of each subject at different ages.}
    \label{fig:LogiFaceDatabase}
    % \vspace{-10mm}
\end{figure}

LogiFace database is collected from various resources.
Celebrity face images were collected in a wide range of ages from $1$ to $99$ with both frontal and profile face images.
Of the annotation of the dataset, we determined the age of person based on the captured time of images or videos. 
The photos were captured as color images with a size of $400 \times 500$ pixels in a PNG format.
The database has $12,656$ full images of $369$ subjects with the average 34 face images per subject.
Comparing to two of the common public longitudinal face databases FG-NET \cite{fgnet_database} and MORPH \cite{ricanek2006morph}, LogiFace is unquie in that it contains more longitudinal subject sequences which are required to capture face images of a subject for a long period.
%unique in that it contains more longitudinal face aging images per subject where requires captures face images for a long time. 
Fig. \ref{fig:LogiFaceDatabase} illustrates examples of our LogiFace database.

% [Sample photos of the LogiFace database]

\begin{table*}[!b]
% 	\footnotesize
	\centering
	\caption{Verification performance (\%) on FR and AIFR datasets, i.e. LFW, CACD-VS, AgeDB, FG-NET (LOO), and FG-NET (MF1). %While the baseline denotes the self-studied student, the Student-1 and Student-2 are the guided students by using $\ell_2$ and the proposed angular distillation losses, respectively.
	%Given a teacher network, Student-1 only learns from teacher's features in the last stage while other students (i.e. Student-2, Student-3, Student-4) are further supervised with the increasing number of intermediate stages.
	} 
	
	\label{tb:SmallScaleBenchmark} 
	%\footnotesize
	\resizebox{1.\textwidth}{!} {
	\begin{tabular}{|c | c c | l | c | c c c c |}
% 		\Xhline{2\arrayrulewidth}
        \hline
		\textbf{Backbone} & \begin{tabular}{@{}c@{}}\textbf{\# of} \\ \textbf{params}\end{tabular} & \textbf{Ratio} & \textbf{Model Type} &
		\begin{tabular}{@{}c@{}}\textbf{LFW}\end{tabular} & \begin{tabular}{@{}c@{}}\textbf{LogiFace}\\ \end{tabular}& \begin{tabular}{@{}c@{}}\textbf{AgeDB}\\ \end{tabular}& \begin{tabular}{@{}c@{}}\textbf{FG-NET}\\ \textbf{(LOO)} \end{tabular}& \begin{tabular}{@{}c@{}}\textbf{FG-NET}\\ \textbf{(MF1)} \end{tabular}\\
		\hline
		%\hline
		\begin{tabular}{@{}l@{}} ResNet90 \cite{he2016deep}\end{tabular} &  63.67M & 100\% &Teacher &   
		                                      99.82\%          & 80.11\%         & 98.37\%          & 96.31\%          & 66.30\% \\
		\hline
		\hline
		\multirow{4}{*}{\begin{tabular}{@{}c@{}}MobileNetV1 \cite{howard2017mobilenets} \\ (MV1)\end{tabular}} & \multirow{4}{*}{3.53M}& \multirow{4}{*}{5.54\%} & Self-Studied  &    
		                                      99.53\%          &  76.97\%         & 96.30\%          & 93.41\%          & 54.06\% \\
		& & & Student-1 ($\ell_2$ loss)&      99.60\%          &  77.12\%         & 96.83\%          & 93.51\%          & 54.66\% \\
		& & & \textbf{MV1-LIAD (Ours)}&                \textbf{99.63\%} & \textbf{77.81\%} & \textbf{97.10\%} & \textbf{93.71\%} & \textbf{56.78\%} \\
		& & & \textbf{MV1-LIAAD (Ours)}&               \textbf{99.67\%} & \textbf{77.81\%} & \textbf{97.13\%} & \textbf{93.81\%} & \textbf{57.20\%}\\
		
		\hline
		\multirow{4}{*}{\begin{tabular}{@{}c@{}}MobileNetV2 \cite{sandler2018mobilenetv2} \\ (MV2)\end{tabular}} & \multirow{4}{*}{2.15M} &  \multirow{4}{*}{3.38\%} & Self-Studied  &  
		                                      99.42\%          & 76.28\%          & 95.28\%          & 92.42\%          & 49.14\% \\
		& & & Student-1 ($\ell_2$ loss)&      99.53\%          & 76.36\%          & 96.20\%          & 93.41\%          & 52.28\% \\
		& & & \textbf{MV2-LIAD (Ours)}&                \textbf{99.63\%} & \textbf{76.89\%} & \textbf{97.00\%} & \textbf{93.01\%} & \textbf{53.64\%} \\
		& & & \textbf{MV2-LIAAD (Ours)}&               \textbf{99.63\%} & \textbf{77.20\%} & \textbf{97.13\%} & \textbf{93.61\%} & \textbf{54.23\%} \\
		
		 \hline
		 \multirow{4}{*}{\begin{tabular}{@{}c@{}}MobileFacenet \cite{chen2018mobilefacenet} \\ (MFN)\end{tabular}} & \multirow{4}{*}{1.2M} & \multirow{4}{*}{1.88\%} &Self-Studied & 
		                                      99.45\%          & 75.49\%          & 96.17\%          & 92.72\%          & 49.84\% \\
		 & & & Student-1 ($\ell_2$ loss)&     99.50\%          & 75.75\%          & 96.45\%          & 92.81\%          & 50.82\% \\
		 & & & \textbf{MFN-LIAD (Ours)} &              \textbf{99.60\%} & \textbf{76.59\%} & \textbf{96.73\%} & \textbf{93.01\%} & \textbf{52.41\%} \\
		 & & & \textbf{MFN-LIAAD (Ours)} &             \textbf{99.63\%} & \textbf{77.20\%} & \textbf{96.83\%} & \textbf{93.11\%} & \textbf{53.46\%} \\
		
		 \hline
		 \multirow{4}{*}{\begin{tabular}{@{}c@{}}MobileFacenet-R \\ (MFNR)\end{tabular}} & \multirow{4}{*}{3.73M} & \multirow{4}{*}{5.86\%}& Self-Studied & 
		                                      99.60\%          &  76.91\%         & 96.90\%          & 93.51\%          & 58.39\% \\
		 & & & Student-1 ($\ell_2$ loss)&     99.68\%          &  77.48\%         & 97.48\%          & 93.91\%          & 58.85\% \\
		 & & & \textbf{MFNR-LIAD (Ours)} &             \textbf{99.77\%} & \textbf{77.65\%} & \textbf{97.63\%} & \textbf{94.21\%} & \textbf{59.81\%} \\
		 & & & \textbf{MFNR-LIAAD (Ours)} &            \textbf{99.75\%} & \textbf{77.81\%} & \textbf{97.73\%} & \textbf{95.11\%} & \textbf{60.11\%}  \\
		\hline
% 		\Xhline{2\arrayrulewidth}
	\end{tabular}
	}
	%\vspace{-5mm}
\end{table*}

\subsection{Implementation Details}
%\noindent
\textbf{Databases.} Our training data includes MS-Celeb-1M \cite{guo2016ms} cleaned by \cite{deng2018arcface} with 5.8M photos from 85K identities for FR learning and IMDB-WIKI \cite{Rothe-ICCVW-2015} for Age Estimation learning. We use  MS-Celeb-1M to train FR teacher $\mathcal{T}^R$ and IMDB-WIKI to train age estimation teacher $\mathcal{T}^A$. Then MS-Celeb-1M is used to train the student $\mathcal{S}$ with the supports of both $\mathcal{T}^R$ and $\mathcal{T}^A$.
For validation, both small-scale and large-scale protocols on AiFR and FR are adopted. 
In particular, with small-scale protocols, we conduct the  experiments on public AiFR datasets including 
AgeDB \cite{moschoglou2017agedb}, FG-NET \cite{fgnet_database}, and our collected LogiFace. While the first two protocols follow the set up of 10-fold validation with positive and negative matching pairs across ages, the latter adopts the leave-one-out (LOO) validation as in \cite{DAL}. For large-scale benchmark, we validate our proposed LIAAD approach on the challenging Megaface Protocol 1 on FG-NET set \cite{kemelmacher2016megaface} against million-scale of distractors. Furthermore, in order to further evaluate the efficiency and robustness of LIAAD approach, the performance on standard FR benchmarks including LFW \cite{huang2008labeled}, Megaface with FaceScrub Set \cite{kemelmacher2016megaface}, IJB-B \cite{whitelam2017iarpa}, and IJB-C \cite{maze2018iarpa} is also reported.

% \afterpage{
% \begin{savenotes}

%\noindent
\textbf{Data Preprocessing.} All faces 
in the training and testing database 
are detected using MTCNN \cite{MTCNN} and aligned to a predefined template using similarity transformation. They are cropped to the size of $112 \times 112$
and pixel values are normalized to $[-1,1]$.

%\noindent
\textbf{Network Architectures.}
In all experiments, ResNet-90 \cite{he2016deep} is used as the teacher network, while different light-weight networks, i.e. MobileNetV1 \cite{howard2017mobilenets}, MobileNetV2 \cite{sandler2018mobilenetv2}, MobileFaceNet \cite{chen2018mobilefacenet}. A modified version of MobileFacenet, namely MobileFacenet-R, is also adopted for the student network. This modified version is similar to MobileFacenet except the feature size of each ResNet-block is equal to the size of its corresponding features in the teacher network.

%\noindent
\textbf{Model Configurations.} 
%We reuse the configuration as in \cite{wang2018cosface, deng2018arcface} where the normalized versions of the features $F^t_n$ and $F^s_n$ are scaled to 64. 
In the training stage, the batch size is set to 512. The learning rate starts from 0.1 and the momentum is 0.9. All the models are trained in MXNET environment with a machine of Core i7-6850K @3.6GHz CPU, 64.00 GB RAM with four P6000 GPUs.
The $\lambda_n$ is experimentally set to 1 in case of Angular Distillation Loss while this parameter is set to 0.001 as the case of $\ell_2$ loss due to the large value of the loss with large feature map. For the intermediate layers, $\lambda_i = \frac{\lambda_{i+1}}{2}$.

\subsection{Evaluation Results}

\begin{table} [!b]
	%\small
	\centering
	\caption{Verification performance (\%) on LFW.} 
	%\footnotesize
	\begin{tabular}{| l| c| c| c|}
        \hline
		\textbf{Method}  & 
		\begin{tabular}{@{}c@{}}\textbf{Training}\textbf{ Data}\end{tabular}& \begin{tabular}{@{}c@{}}\textbf{Structure Size}\\ \end{tabular} &\begin{tabular}{@{}c@{}}\textbf{Accuracy}\\ \end{tabular} \\
		\hline
		%\hline
		Deep-ID2+ (NIPS, 2014) \cite{sun2014deep} &  0.3M &  Large& 99.47\%\\
		FaceNet (CVPR, 2015) \cite{schroff2015facenet} &  200.0M &  Large& 99.63\%\\
		Center Loss (ECCV, 2016) \cite{wen2016discriminative} &  0.7M&  Large& 99.28\%\\
		Sphereface (CVPR, 2017) \cite{liu2017sphereface} &  0.5M &  Large& 99.42\% \\
		RangeLoss (CVPR, 2017) \cite{zhang2017range} &  5.0M &  Large& 99.52\%\\
		Sphereface+ (NIPS, 2018) \cite{liu2018learning} &  0.5M &  Large& 99.47\% \\
		Marginal Loss (CVPR, 2018) \cite{Wang_2018_CVPR}&  4.0M &  Large& 99.48\%\\
		CosFace (CVPR, 2019) \cite{Wang_2018_CVPR}&  5.0M &  Large& 99.73\%\\
		ArcFace (CVPR, 2019) \cite{deng2018arcface} &  5.8M &  Large & 99.83\% \\
        MFFD (ACM MM, 2020) \cite{meng2020multifeat} &  0.7M & Large & 97.65\%\\
        MTLFace (CVPR, 2021) \cite{huang2020mtlface} &  5.0M & Large & 99.52\% \\
		LightCNN (TIFS, 2018) \cite{wu2018light}&  4.0M & Small  &99.33\%\\
		Li-ArcFace (ICCV, 2019) \cite{Li_2019_ICCV_Workshops} &  5.8M & Small & 99.27\% \\
		VarGFaceNet (ICCV, 2019) \cite{Yan_2019_ICCV_Workshops} & 5.8M & Small & 99.68\% \\
		\hline %\hline
		\textbf{MV1-LIAAD (Ours)} &  5.8M &  Small & \textbf{99.63\%}\\
		\textbf{MV2-LIAAD (Ours)} &  5.8M &  Small & \textbf{99.63\%}\\
		\textbf{MFN-LIAAD (Ours)} &  5.8M &  Small & \textbf{99.60\%}\\
		\textbf{MFNR-LIAAD (Ours)} & 5.8M &  Small & \textbf{99.77\%}\\
		\hline
% 		{ViT-L} (Teacher) & 5.8M &  Large & 99.85\%\\
		{ViT-T with $l_2$} & 5.8M &  Small & 99.68\%\\
		\textbf{ViT-T-LIAD (Ours)}  & 5.8M &  Small &\textbf{ 99.80\%}\\
		\hline
	\end{tabular}\label{tb:LFWBenchmark}
	%\vspace{-6mm}
\end{table}

\begin{table}[!b]
	%\small
	\centering
	\caption{Comparison with other methods on AIFR benchmarks.} 
	%\footnotesize
	\begin{tabular}{|l|c|c c|}
		\hline
		\textbf{Method} & \begin{tabular}{@{}c@{}}\textbf{Structure}\\\textbf{Size}\end{tabular} & \begin{tabular}{@{}c@{}}\textbf{FG-NET} \\\textbf{(LOO)} \end{tabular}& \begin{tabular}{@{}c@{}}\textbf{FG-NET}\\ \textbf{(MF1)} \end{tabular}\\
	   % \hline
	    \hline
	    %HD-LBP \cite{HD_LBP} & & 81.6\% & $-$ & $-$\\
	    HFA (ICCV, 2013) \cite{HFA} & $-$ &   69.00\% &$-$\\
	    MEFA (CVPR, 2015) \cite{MEFA} & $-$ &   76.20\% &$-$\\
	    CAN (NeuComp, 2017) \cite{CAN} & $-$ &   86.50\% &$-$\\
	    LF-CNNs (CVPR, 2016) \cite{LFCNNs} & Large & 88.10\% &$-$\\
	    A-Softmax (ISCSLP, 2018) \cite{ASoftmax} &Large  & $-$ & 46.77\%\\
	    OE-CNNs (ECCV, 2018) \cite{OECNN} & Large & $-$ &52.67\%\\
	    DAL (CVPR, 2019) \cite{DAL} & Large & 94.50\% & 57.92\%\\
	    MTLFace (CVPR, 2021) \cite{huang2020mtlface} & Large & 94.78\% & 57.92\% \\
	    AIM (TPAMI, 2022) \cite{Zhao_PAMI2022} & Large & 93.20\% & 60.94\% \\
	    \hline
	    \textbf{MV1-LIAAD (Ours)} & Small& \textbf{93.71\%} & \textbf{57.20\%}\\
	    \textbf{MV2-LIAAD (Ours)} & Small & \textbf{93.61\%} & \textbf{54.23\%}\\
	    \textbf{MFN-LIAAD (Ours)} & Small & \textbf{93.11\%} & \textbf{53.46\%}\\
	    \textbf{MFNR-LIAAD (Ours)} & Small & \textbf{95.11}\% & \textbf{60.11\%}\\
        \hline
	\end{tabular}\label{table:BlackboxDifferentMatcher}
	%\vspace{-6mm}
\end{table}

%\noindent
\textbf{AiFR Protocols.} 
We validate the efficiency of our LIAAD framework with four light-weight backbones on different protocols including LFW, LogiFace (LOO), AgeDB, FG-NET (LOO) and FG-NET (MF1). The ResNet-90 trained on MS-Celeb-1M acts as the teacher network. Meanwhile, MobileNetV1 (MV1), MobileNetV2 (MV2), MobileFaceNet (MFN), and MobileFacenet-R (MFNR) are chosen as the student networks.
Then, for each light-weight backbone, four cases are considered: (1) Self-studied student trained without the help from teacher; (2) Student-1 trained using the objective function as Eqn. (\ref{eqn:OverallLoss}) but only $\ell_2$ function is adopted for distillation loss; and our proposed framework with (3) Lightweight Angular Distillation (LIAD) and (4) Lightweight Attentive Angular Distillation (LIAAD).

Table \ref{tb:SmallScaleBenchmark} illustrates the performance of the teacher network together with its students. Rank-1 accuracy is reported for LOO validations.
Due to the limited-capacity of the light-weight backbones, in all four cases, the self-studied networks leave the performance gaps of 0.22\% $-$ 0.40\%, 3.14\% $-$ 4.62\%, 1.47\% $-$ 3.09\%, and 2.80\% $-$ 3.89\% with their teacher on LFW, LogiFace, AgeDB, and FG-NET (LOO) respectively. 
The guided students using $\ell_2$ loss can slightly improve the accuracy. However, we notice that the training process with $\ell_2$ loss is unstable.
%the performance gain is still limited. 
Meanwhile, our proposed framework with LIAD framework efficiently distills the knowledge from the teacher to its student with
%with ``softer'' constraints as well as gives the student the flexibility of interpret the information. 
%more ``place to play and interpret'' the knowledge. 
%As a result, 
the best performance gaps that are significantly reduced to only 0.05\%, 2.30\%, 0.74\%, and 2.10\% on the four benchmarks, respectively. 
Moreover, by adopting LIAAD framework, although the accuracy on LFW keeps comparable to AD ones, the performances on aging benchmarks are reduced to only 0.07\%, 1.91\%, 0.64\%, and 1.20\% on these benchmarks. 
%From these results in Table \ref{tb:LFWBenchmark}, even with the light-weight backbone, our AAD can achieve competitive performance with other large-scale networks.
%

On the MegaFace benchmark with FG-NET (MF1) probe set, 
%the self-studied networks of MV1, MV2, MFN and MFRN stand at 54.06\%, 49.14\%, 49.84\%, and 58.39\%, respectively.
compared to the self-studied students, 
the ones with LIAD improve by 1.42\% to 4.5\% according to the capacity of the students.
%2.72\%, 4.50\%, 2.57\%, and 1.42\%, respectively.
Then, by apdoting LIAAD, these improvements are further made to 1.72\% - 5.09\%.
%the AAD framework helps to further improve the performance on the top the self-studied networks by 3.14\%, 5.09\%, 3.62\%, and 1.72\%, respectively. 
The comparisons with different methods on  FG-NET (LOO), FG-NET (MF1), and LFW are also reported in Tables \ref{tb:LFWBenchmark} and Table \ref{table:BlackboxDifferentMatcher}. These results emphasize the advantages of our LIAAD by outperforming DAL\cite{DAL} even when  the light-weight structure MFNR is adopted.
%From these results, one can see that even with the light-weight structure, MFNR-AAD can outperforms DAL\cite{DAL} 
%
%As shown in Table \ref{table:BlackboxDifferentMatcher}, our methods are compared against to HD-LBP \cite{HD_LBP}, HFA \cite{HFA}, MEFA \cite{MEFA}, CAN \cite{CAN}, LF-CNNs \cite{LFCNNs}, A-Softmax \cite{ASoftmax}, OE-CNNs \cite{OECNN}, DAL \cite{DAL} on FG-NET (LOO) and FG-NET (Megaface) datasets.
%Comparing to A-Softmax, OE-CNNS, DAL where the structures are large, our proposed MV1-ADD, MV2-ADD, MFN-ADD, MFNR-ADD are of small structures while keeping the SOTA results on the benchmark datasets.

% \textcolor{blue}{
Additionally, Table \ref{tb:LFWBenchmark} has illustrated the flexibility of our proposed method. In particular, we have conducted experiments with ResNet-based backbones (i.e. MV1, MV2, MFN, and MFNR) and Transformer-based backbones \cite{vit_paper} (i.e. ViT-S). In the Transformer-based backbone experiments, ViT-L has been used as the teacher network, which has achieved a verification performance on LFW of 99.85\%. 
The experimental results have shown that our proposed losses have consistently improved the performance of various backbones of student networks.
% }

	\begin{table}[!t]
    	\caption{Comparison with different methods on Megaface protocol.} 
    	\label{tb:MegafaceBenchmark} 
    	\centering
    % 	\footnotesize
    % 	\small
    % 	\resizebox{1.0\textwidth}{!} {
    	\begin{tabular}{ >{\arraybackslash}l c c c}
    % 		\Xhline{2\arrayrulewidth}
            \hline
    		\textbf{Method}  &
    		\begin{tabular}{@{}c@{}}\textbf{Protocol}\end{tabular}& \begin{tabular}{@{}l@{}}\textbf{Structure Size}\\ \end{tabular}& \begin{tabular}{@{}l@{}}\textbf{Accuracy}\\ \end{tabular} \\
    		\hline
    		%\hline
    		Sphereface (CVPR, 2017) \cite{liu2017sphereface} & Small & Large& 72.73\% \space \space \\
    		Sphereface+ (NIPS, 2018) \cite{liu2018learning} & Small & Large& 73.03\% \space \space \\
    		%Center Loss \cite{wen2016discriminative} & Small& Large&  65.49\% \space \space\\
    		\hline
    		FaceNet (CVPR, 2015) \cite{schroff2015facenet} & Large & Large& 70.49\% \space \space\\
    		CosFace (CVPR, 2019) \cite{Wang_2018_CVPR}& Large & Large& 82.72\% \space \space\\
    % 		ArcFace \cite{deng2018arcface} & Large & 81.03\%\\
    % 		ArcFace \cite{deng2018arcface} & Large & Large& 98.35\%\footnote{refers to the accuracy obtained by using the refined testing dataset with cleaned labels from \cite{deng2018arcface}.}\\
    		ArcFace-ResNet34 (CVPR, 2019) \cite{deng2018arcface} & Large & Large& 96.70\%\\
    % 		\footnotemark\\
    % 		\footnotetext{Refers to the accuracy obtained by using the refined testing dataset with cleaned labels from \cite{deng2018arcface}.}
    		\hline %\hline
    		MV1 \cite{howard2017mobilenets} & Large & Small& 91.93\% \\% \footnotemark[\value{footnote}] \\ %$^*$\\
    		\textbf{MV1-LIAAD (Ours)} & Large & Small &\textbf{94.16\%} \\% \footnotemark[\value{footnote}] \\ %$^*$\\
    		\hline
    		MV2 \cite{sandler2018mobilenetv2} & Large & Small & 89.22\% \\% \footnotemark[\value{footnote}] \\ %%$^*$\\
    		\textbf{MV2-LIAAD (Ours)} & Large & Small & \textbf{92.86\%} \\% \footnotemark[\value{footnote}] \\ %$^*$\\
    		\hline
    		MFN \cite{chen2018mobilefacenet} & Large & Small& 89.32\% \\% \footnotemark[\value{footnote}] \\ %$^*$\\
    		\textbf{MFN-LIAAD (Ours)} & Large & Small& \textbf{91.89\%} \\% \footnotemark[\value{footnote}] \\ %$^*$\\
    		\hline
    		MFNR & Large & Small & 93.74\% \\% \%\footnotemark[\value{footnote}] \\ %$^*$\\
    		\textbf{MFNR-LIAAD (Ours)} & Large & Small& \textbf{ 95.64\%} \\% \footnotemark[\value{footnote}] \\ %$^*$\\
    % 		\Xhline{2\arrayrulewidth}
            \hline
    	\end{tabular}
    % }
% 	\end{subtable}
`   \end{table}

\noindent
\textbf{Megaface Protocol.} We further evaluate our LIAAD approach on the challenging Megaface benchmark against millions of distractors on both FaceScrub and FG-NET sets. The comparison in terms of the Rank-1 identification rates between our LIAAD and other models is presented in Table \ref{tb:MegafaceBenchmark} for FaceScrub set and last column of Table  \ref{tb:SmallScaleBenchmark} for FG-NET set. These results again show the advantages of our LIAAD with consistent improvements provided to the four light-weight student networks. 
Particularly, with FaceScub probe set, the performance gains for the MV1, MV2, MFN, and MFNR are 2.23\%, 3.64\%, 2.57\%, and 1.9\%, respectively\footnote{Refers to the accuracy obtained by using the refined testing dataset with cleaned labels from \cite{deng2018arcface}.}. Moreover, the MFNR-LIAAD achieves competitive accuracy (i.e. 95.64\%) with other large-scale networks and reduces the gap with ArcFace \cite{deng2018arcface} to only 1.06\%.

	\begin{table}[!t]
    % 	\footnotesize
    	\caption{Comparison with different methods on 1:1 IJB-B and IJB-C. The accuracy is reported at TAR (@FAR=1e-4).} 
    	\label{tb:IJBBenchmark} 
    	\centering
    	\resizebox{1.0\textwidth}{!} {
    	\begin{tabular}{l c c c}
            \hline
    		\textbf{Method}  & \begin{tabular}{@{}c@{}}\textbf{Training}  \textbf{Data}\end{tabular} &
    		\begin{tabular}{@{}c@{}}\textbf{IJB-B}\end{tabular}& \begin{tabular}{@{}c@{}}\textbf{IJB-C}\\ \end{tabular} \\
    		\hline
    		%\hline
    		SENet50 (FG, 2018) \cite{cao2018vggface2} & VGG2 & 0.800 & 0.840 \\
    		MN-VC (BMVC, 2018) \cite{xie2018multicolumn} & VGG2 & 0.831 & 0.862\\
    		ResNet50 + DCN (ECCV, 2018) \cite{Xie18a} & VGG2 & 0.841 & 0.880\\
    		ArcFace (CVPR, 2019) \cite{deng2018arcface} & VGG2& 0.898 & 0.921\\
    		ArcFace (CVPR, 2019) \cite{deng2018arcface} & MS1M & 0.942 & 0.956\\
    		\hline %\hline
    		MV1 \cite{howard2017mobilenets} & MS1M & 0.909 &0.930\\
    		\textbf{MV1-LIAAD (Ours)} & MS1M & \textbf{0.915} &\textbf{0.936}\\
    		\hline
    		MV2 \cite{sandler2018mobilenetv2} & MS1M & 0.883 &0.907\\
    		\textbf{MV2-LIAAD (Ours)} & MS1M & \textbf{0.902} &\textbf{0.922}\\
    		\hline
    		MFN \cite{chen2018mobilefacenet} & MS1M & 0.896&0.918\\
    		\textbf{MFN-LIAAD (Ours)} & MS1M & \textbf{0.898} &\textbf{0.921}\\
    		\hline
    		MFNR & MS1M & 0.912& 0.916\\
    		\textbf{MFNR-LIAAD (Ours)} & MS1M & \textbf{0.923} &\textbf{0.940}\\
            \hline
    	\end{tabular}
    	}
% 	\end{subtable}
    \end{table}
% \end{table}

% \footnotetext{Refers to the accuracy obtained by using the refined testing dataset with cleaned labels from \cite{deng2018arcface}.}

%\noindent
\textbf{IJB-B and IJB-C Protocols.} The comparisons against other recent methods on IJB-B and IJB-C benchmarks are also illustrated in Table \ref{tb:IJBBenchmark}. Similar to the Megaface protocols, our LIAAD is able to boost the performance of the light-weight backbones significantly and reduces the performance gap to the large-scale backbone to only 0.019 on IJB-B and 0.016 on IJB-C. 
%Figure XXX presents the ROC curves of ShrinkTeaNet and its teacher on both IJB-B and IBJ-C benchmarks. 
These results have further emphasized the advantages of the proposed LIAAD for model distillation.

% \textcolor{blue}{
\textbf{Computational Consumption}
To illustrate the performance of the student's networks in the term of computation, we report the inference FLOPS of the teacher and student networks. As shown in Table \ref{tab:inference_flops}, the inference FLOPs of the student networks, i.e. MV1, MV2, MFN, MFNR and ViT-T, are significantly smaller than the teacher networks, i.e. ResNet 90, ViT-L. However, these students network have achieved the relatively state-of-the-art performance compared to the teacher network as aforementioned.
% }


\begin{table}[t]
\centering
\caption{Computational FLOPs of Different Backbone Networks.}
\label{tab:inference_flops}
\begin{tabular}{c|c|c}
    \textbf{ Method}          & \textbf{Model Type}  & \textbf{Inference FLOPS}         \\ 
    \hline
    \hline
    ResNet 90              & Teacher     & 12.122 GFLOPs \\ 
    MV1                      & Student     & 0.575 GFLOPS \\
    MV2                      & Student     & 0.300 GFLOPS \\
    MFN                    & Student     & 0.449 GFLOPs  \\ 
    MFNR                   & Student     & 1.845 GFLOPs  \\ 
    \hline
    \hline
    ViT-L                  & Teacher     & 24.584 GFLOPs \\ 
    ViT-T                  & Student     & 1.379 GFLOPs \\
\end{tabular}
\end{table}

\section{Discussions}

\noindent
% \textcolor{blue}{
\textbf{Advantages:} Our work presents a novel approach to face-invariant face recognition and gains superiority compared to prior approaches. First, our approach takes advantages of both face recognition datasets with and without age labels. With our proposed distillation framework, our approach is able to improve the age-invariant property and performance of the face recognition models. Second, our approach gives a flexibility in choosing network backbone. In particular, our proposed framework can adopt any network backbones to and still be able to improve the performance of face recognition. Third, our approach does not increase the computational cost. Different with several prior approaches (e.g. VarGFaceNet \cite{Yan_2019_ICCV_Workshops}), our focus does not lying on modifying the network backbones. We aim to provide a well-designed angular distillation framework to improve the performance of the face recognition models without increasing the computational cost of the inference procedure. The intensive experiments in the previous section have shown the superior performance of our approach compared to prior methods. 
% }

\noindent
% \textcolor{blue}{
\textbf{Limitations and Future Works:}
In our work, the teacher networks still require supervised training. In particular, training the two teacher network requires labeled datasets. Hence, our approach has not taken advantage of the large-scale unlabeled datasets yet. 
This could remain a weak point in our approach. Additionally, in our approach, we have focus on the effect of age-invariant features to the face recognition models. However, in practice, there are many other factors, e.g. emotion, ethnicity, gender, etc, could affect the performance of the face recognition models. These factors have not been well handled by our approach yet. In future works, we will further exploit unsupervised and self-supervised approaches to leverage the unlabeled large-scale datasets. Also, we take other facial factors into account to improve the performance of the face recognition engines.
% }

\section{Conclusions}

% \textcolor{blue}{
This paper has presented the novel Lighweight Attentive Angular Distillation paradigm for age-invariant open-set face recognition.
Our approach aims to strengthen the lightweight student networks as powerful as their teachers. 
Particularly, by adopting the proposed Age-invariant Attentive and Angular Distillation Losses for the distillation of feature embedding process, the student network can absorb the knowledge of the teacher's hypersphere and age-invariant attention in an efficient manners. 
These learned knowledges can be flexibly adopted even when the testing classes are different from the training ones. The intensive evaluation in both small-scale and large-scale protocols have showed the advantages of the proposed LIAAD framework.
% }

%
% \textcolor{blue}{
To the best of our knowledge, this work has introduced one of the first studies to utilize the advantages of both real-world datasets: (i) face recognition datasets, i.e. large-scale but without age labels, and (ii) face aging datasets, i.e. with age labels but not many samples per subject. In fact, our approach is far apart from prior AiFR methods that require age-labeled FR training data which is usually limited in practice. Also, prior works are unable to exploit the knowledge from standard face recognition datasets, i.e. the age-annotation is not exploited.
Besides, our work has provided a new point of view to approach the AiFR problem. 
Indeed, our approach is more efficient, more tractable and data reusable in the performance improvement thanks to the proposed distillations compared against other approaches.
In addition, in the discovery of this work, as shown in Fig. \ref{fig:Peri_heatmap}, age-invariant attention region is automatically focused on the face's periocular regions which is consistent with the finding from \cite{periocular_KhoaLuu}. 
% }

\noindent
\textbf{Acknowledgement} This work is supported by NSF Data Science, Data Analytics that are Robust and Trusted (DART), Arkansas Biosciences Institute (ABI) Grant Program, and NSF WVAR-CRESH Grant. 

% \section*{References}

\bibliography{mybibfile}

\end{document}